\documentclass[aoas,nameyear,dvips]{arximspdf}
\usepackage{mathbh}
\usepackage{graphicx}

\doi{10.1214/09-AOAS308}
\volume{4}
\issue{1}
\pubyear{2010}
\firstpage{94}
\lastpage{123}

\makeatletter
\renewcommand{\citep}[1]{[\citet{#1}]}
\newcommand{\eqref}[1]{(\ref{#1})}

\newcommand{\sign}{\operatorname{sign}}
\newcommand{\argmin}{\mathop{\arg\min}}
\newcommand{\TV}{\operatorname{TV}}
\newcommand{\norm}[1]{\|#1\|}
\newcommand{\ind}{\mathbh{1}}
\newcommand{\xv}{\mathbf{x}}
\newcommand{\Xv}{\mathbf{X}}
\newcommand{\thetav}{\bolds{\theta}}
\newcommand{\Sigmav}{\bolds{\Sigma}}
\newtheorem{theorem}{Theorem}
\makeatother

\begin{document}
\begin{frontmatter}

\title{Estimating time-varying networks}
\runtitle{Estimating time-varying networks}

\begin{aug}
\author[a]{\fnms{Mladen} \snm{Kolar}\ead[label=e1]{mladenk@cs.cmu.edu}},
\author[a]{\fnms{Le} \snm{Song}\thanksref{tz}\ead[label=e2]{lesong@cs.cmu.edu}},
\author[a]{\fnms{Amr} \snm{Ahmed}\ead[label=e3]{amahmed@cs.cmu.edu}} \and
\author[a]{\fnms{Eric P.} \snm{Xing}\corref{}\thanksref{tg}\ead[label=e4]{epxing@cs.cmu.edu}}
\runauthor{Kolar, Song, Ahmed and Xing}
\affiliation{Carnegie Mellon University}
\thankstext{tz}{Supported by a Ray and Stephenie Lane Research Fellowship.}
\thankstext{tg}{Supported by Grant ONR N000140910758, NSF DBI-0640543, NSF DBI-0546594, NSF IIS-0713379 and an
Alfred P. Sloan Research Fellowship.  Corresponding author.}
\address[a]{School of computer science\\
Carnegie Mellon university\\
5000 Forbes Ave\\
8101 Gates-Hillman Center\\
Pittsburgh, Pennsylvania 15213\\
USA\\
\printead{e1}\\
\phantom{E-mail:\ }\printead*{e2}\\
\phantom{E-mail:\ }\printead*{e3}\\
\phantom{E-mail:\ }\printead*{e4}} 
\end{aug}

\received{\smonth{12} \syear{2008}}
\revised{\smonth{8} \syear{2009}}

\begin{abstract}
Stochastic networks are a plausible representation of the relational
information among entities in dynamic systems such as living cells
or social communities. While there is a rich literature in
estimating a static or temporally invariant network from observation
data, little has been done toward estimating time-varying networks
from time series of entity attributes. In this paper we present two
new machine learning methods for estimating time-varying networks,
which both build on a temporally smoothed $l_1$-regularized logistic
regression formalism that can be cast as a standard
convex-optimization problem and solved efficiently using generic
solvers scalable to large networks. We report promising results on
recovering simulated time-varying networks. For real data sets, we
reverse engineer the latent sequence of temporally rewiring
political networks between Senators from the US Senate voting
records and the latent evolving regulatory networks underlying 588
genes across the life cycle of \textit{Drosophila melanogaster} from
the  microarray time course.
\end{abstract}


\end{frontmatter}

\setcounter{footnote}{2}
\section{Introduction}

In many problems arising from natural, social, and information
sciences, it is often necessary to analyze a large quantity of random
variables interconnected by a complex dependency network, such as the
expressions of genes in a genome, or the activities of individuals in
a community.  Real-time analysis of such networks is important for
understanding and predicting the organizational processes, modeling
information diffusion, detecting vulnerability, and assessing the
potential impact of interventions in various natural and built
systems. It is not unusual for network data to be large, dynamic,
heterogeneous, noisy, incomplete, or even unobservable. Each of these
characteristics adds a degree of complexity to the interpretation and
analysis of networks. In this paper we present a new methodology and
analysis that address a particular aspect of dynamic network analysis:
how can one reverse engineer networks that are latent, and
topologically evolving over time, from time series of nodal
attributes.

While there is a rich and growing literature on modeling
time-invariant networks, much less has been done toward modeling
dynamic networks that are rewiring over time. We refer to these time
or condition specific circuitries as \textit{time-varying networks},
which are ubiquitous in various complex systems. Consider the
following two real world problems:
\begin{itemize}
\item \textit{Inferring gene regulatory networks.} Over the course of
organismal development, there may exist multiple biological
``themes'' that dynamically determine the functions of each gene
and their regulations. As a result, the regulatory networks at
each time point are context-dependent and can undergo systematic
rewiring rather than being invariant over
time~\citep{LusBabYuEtal2004}. An intriguing and unsolved problem
facing biologists is as follows: given a set of microarray measurements
over the expression levels of $p$ genes, obtained at $n$ ($n \ll
p$) different time points during the developmental stages of an
organism, how do you reverse engineer the time-varying regulatory
circuitry among genes?
\item \textit{Understanding stock market.} In a finance setting we have
values of different stocks at each time point. Suppose, for
simplicity, that we only measure whether the value of a particular
stock is going up or down. We would like to find the underlying
transient relational patterns between different stocks from these
measurements and get insight into how these patterns change over
time.
\end{itemize}

In both of the above-described problems, the data-generating process
and latent relational structure between a set of entities change over
time. A key technical hurdle preventing us from an in-depth
investigation of the mechanisms underlying these complex systems is
the unavailability of \textit{serial snapshots} of the time-varying
networks underlying these systems. For example, for a realistic
biological system, it is impossible to experimentally determine
time-specific networks for a series of time points based on current
technologies such as two-hybrid or ChIP-chip systems.  Usually, only
time series measurements, such as microarray, stock price, etc., of
the activity of the nodal entities, but not their linkage status, are
available. Our goal is to recover the latent time-varying networks
with temporal resolution up to every single time point based on time
series measurements. Most of the existing work on structure estimation
assumes that the data generating process is time-invariant and that
the relational structure is fixed, which may not be a suitable
assumption for the described problems. The focus of this paper is to
estimate dynamic network structure from a time series of entity
attributes.

The Markov Random Fields (MRF) have been a widely studied model for
the relational structure over a fixed set of entities
[\citet{wainwright08graphical}; \citet{getoor07introduction}]. Let $G = (V, E)$
be a graph with the vertex set $V$ and the edge set $E$. A node $u \in
V$ represents an entity (e.g., a stock, a gene, or a person) and an
edge $(u, v) \in E$ represents a relationship (e.g., correlation,
friendship, or influence). Each node in the vertex set $V = \{1,
\ldots, p\}$ corresponds to an element of a $p$-dimensional random
vector $\Xv = (X_1, \ldots, X_p)'$ of nodal states, whose probability
distribution is indexed by $\thetav \in \Theta$. Under a MRF, the
nodal states are assumed to be discrete, that is, $\Xv \in \mathcal{X}^p
\equiv \{s_1, \ldots, s_k \}^p$, and the edge set $E \subseteq V
\times V$ encodes conditional independence assumptions among
components of the random vector $\Xv$, more specifically, $X_u$ is
conditionally independent of $X_v$ given the rest of the variables if
$(u,v) \notin E$. In this paper we will analyze a special kind of
MRF in which the nodal states are binary, that is, $\mathcal{X} \equiv
\{-1, 1\}$, and the interactions between nodes are given by pairwise
potentials $\theta_{uv}$ for all $(u, v) \in E$ and $\theta_{uv} = 0$
otherwise. This type of MRF is known as the Ising model, under which
the joint probability of $\Xv = \xv$ can be expressed as a simple
exponential family model: $\mathbb{P}_{\thetav}(\xv) = \frac{1}{Z}\exp
\{\sum_{u < v} \theta_{uv}x_ux_v \}$, where $Z$ denotes the partition
function. Some recent work [\citet{bresler07reconstruction}; \citet{ravikumar09high}] has analyzed the graph structure estimation from
data that are assumed to be the i.i.d. sample from the Ising
model. A particular emphasis was put on \textit{sparsistent} estimation,
that is, consistent estimation of the graph structure, under a setting in
which the number of nodes $p$ in the graph is larger than the sample
size $n$, but the number of neighbors of each node is small, that is, the
true graph is sparse \citep{ravikumar09high}.

In this paper we concern ourselves with estimating the time-varying
graph structures of MRFs from a time series of nodal states
$\{\xv^t\}_{t \in \mathcal{T}_n}$, with $\mathcal{T}_n = \{ 1/n, 2/n,
\ldots, 1\}$ being the time index set, that are independent (but not
identically distributed) samples from a series of time-evolving
MRFs\break
$\{\mathbb{P}_{\thetav^t}(\cdot)\}_{t \in \mathcal{T}_n}$.  This is a
much more challenging and more realistic scenario than the one that
assumes that the nodal states are sampled i.i.d. from a
time-invariant MRF. Our goal is to estimate a sequence of graphs
$\{G^t\}_{t \in \mathcal{T}_n}$ corresponding to observations $\xv^t \sim
\mathbb{P}_{\thetav^t}$ in the time series. The problem of dynamic
structure estimation is of high importance in domains that lack prior
knowledge or measurement techniques about the interactions between
different actors; and such estimates can provide desirable information
about the details of relational changes in a complex system. It might
seem that the problem is ill-defined, since for any time point we have
at most one observation; however, as we will show shortly, under a set
of suitable assumptions the problem is indeed well defined and the
series of underlying graph structures can be estimated. For example,
we may assume that the probability distributions are changing \textit{
smoothly} over time, or there exists a partition of the interval $[0,
1]$ into segments where the graph structure within each segment is
invariant.

\subsection{Related work}
A large body of literature has focused on estimation of the
time-invariant graph structure from the i.i.d. sample. Assume
that $\mathcal{D}_n = \{\xv^{i} = (x_1^{i}, \ldots,
x_p^{i})\}_{i=1}^n$ are $n$ i.i.d. samples from
$\mathbb{P}_{\thetav}$. Furthermore, under the assumption that
$\mathbb{P}_{\thetav}$ is a multivariate normal distribution with mean
vector $\bolds{\mu}$ and covariance matrix $\Sigmav$, estimation of
the graph structure is equivalent to the estimation of zeros in the
concentration matrix $\bolds{\Omega} \equiv \Sigmav^{-1}$
\citep{lauritzen96graphical}. \citet{Drton04model} proposed a method
that tests if partial correlations are different from zero, which can
be applied when the number of dimensions $p$ is small in comparison to
the sample size $n$. In the recent years, research has been directed
toward methods that can handle data sets with relatively few
high-dimensional samples, which are common if a number of domains,
for example, microarray measurement experiments, fMRI data sets, and
astronomical measurements. These ``large $p$, small $n$'' data sets
pose a difficult estimation problem, but under the assumption that the
underlying graph structure is sparse, several methods can be employed
successfully for structure recovery. \citet{meinshausen06high} proposed
a procedure based on \textit{neighborhood selection} of each node via the
$\ell_1$ penalized regression. This procedure uses a
pseudo-likelihood, which decomposes across different nodes, to
estimate graph edges and, although the estimated parameters are not
consistent, the procedure recovers the graph structure consistently
under a set of suitable conditions. A related approach is proposed in
\citet{peng08partial} who consider a different neighborhood selection
procedure for the structure estimation in which they estimate all
neighborhoods jointly and as a result obtain a global estimate of the
graph structure that empirically improves the performance on a number
of networks. These neighborhood selection procedures are suitable for
large-scale problems due to availability of fast solvers to $\ell_1$
penalized problems [\citet{efron04lars}; \citet{friedman07pathwise}].

Another popular approach to the graph structure estimation is the
$\ell_1$ penalized likelihood maximization, which simultaneously
estimates the graph structure and the elements of the covariance
matrix, however, at a price of computational efficiency. The penalized
likelihood approach involves solving a semidefinite program (SDP) and
a number of authors have worked on efficient solvers that exploit the
special structure of the problem
[\citet{banerjee07model}; \citet{yuan07model};
\citet{Friedman07glasso}; \citet{Duchi08projected}; \citet{rothman08spice}]. Of these
methods, it seems that the graphical lasso \citep{Friedman07glasso} is
the most computationally efficient. Some authors have proposed to use
a nonconcave penalty instead of the $\ell_1$ penalty, which tries to
remedy the bias that the $\ell_1$ penalty introduces
[\citet{fan01variable}; \citet{fan08network}; \citet{zou08onestep}].

When the random variable $\Xv$ is discrete, the problem of structure
estimation becomes even more difficult since the likelihood cannot be
optimized efficiently due to the intractability of evaluation of the
log-partition function.  \citet{ravikumar09high} use a
pseudo-likelihood approach, based on the local conditional likelihood
at each node, to estimate the neighborhood of each node, and show that
this procedure estimates the graph structure consistently.

All of the aforementioned work analyzes estimation of a time-invariant
graph structure from an i.i.d. sample. On the other hand, with
few exceptions~[\citet{xingSNA06}; \citet{sarkarmoore2006}; \citet{guohannekefuxing2007}; \citet{zhou08time}],
much less has been done on
modeling dynamical processes that guide topological rewiring and
semantic evolution of networks over time. In particular, very little
has been done toward estimating the time-varying graph topologies
from observed nodal states, which represent attributes of entities
forming a network. \citet{xingSNA06} introduced a new class of models
to capture dynamics of networks evolving over discrete time steps,
called \textit{temporal Exponential Random Graph Models} (tERGMs). This
class of models uses a number of statistics defined on time-adjacent
graphs, for example, ``edge-stability,'' ``reciprocity,'' ``density,''
``transitivity,'' etc., to construct a log-linear graph transition
model $P(G^t|G^{t-1})$ that captures dynamics of topological
changes. \citet{guohannekefuxing2007} incorporate a hidden Markov
process into the tERGMs, which imposes stochastic constraints on
topological changes in graphs, and, in principle, show how to infer a
time-specific graph structure from the posterior distribution of
$G^t$, given the time series of node attributes. Unfortunately, even
though this class of model is very expressive, the sampling algorithm
for posterior inference scales only to small graphs with tens of
nodes.

The work of \citet{zhou08time} is the most relevant to our work and we
briefly describe it below. The authors develop a nonparametric method
for estimation of a time-varying Gaussian graphical model, under the
assumption that the observations $\xv^t \sim \mathcal{N}(0, \Sigmav^t)$
are independent, but not identically distributed, realizations of a
multivariate distribution whose covariance matrix changes smoothly
over time. The time-varying Gaussian graphical model is a continuous
counterpart of the discrete Ising model considered in this paper. In
\citet{zhou08time}, the authors address the issue of consistent, in the
Frobenius norm, estimation of the covariance and concentration matrix,
however, the problem of consistent estimation of the nonzero pattern
in the concentration matrix, which corresponds to the graph structure
estimation, is not addressed there. Note that the consistency of the
graph structure recovery does not immediately follow from the
consistency of the concentration matrix.

The paper is organized as follows. In Section~\ref{sec:methods} we
describe the proposed models for estimation of the time-varying
graphical structures and the algorithms for obtaining the estimators.
In Section~\ref{sec:simulation} the performance of the methods is
demonstrated through simulation studies. In Section
\ref{sec:experiments} the methods are applied to some real world data
sets. In Section~\ref{sec:theory} we discuss some theoretical
properties of the algorithms, however, the details are left for a
separate paper. Discussion is given in Section~\ref{sec:discussion}.

\section{Methods} \label{sec:methods}

Let $\mathcal{D}_n = \{ \xv^t \sim \mathbb{P}_{\thetav^t} \mid t \in
\mathcal{T}_n\}$ be an independent sample of $n$ observation from a time
series, obtained at discrete time steps indexed by $\mathcal{T}_n = \{
1/n, 2/n, \ldots, 1\}$ (for simplicity, we assume that the observations
are equidistant in time). Each sample point comes from a different
discrete time step and is distributed according to a distribution
$\mathbb{P}_{\thetav^t}$ indexed by $\thetav^t \in \Theta$. In
particular, we will assume that $\Xv^t$ is a $p$-dimensional random variable
taking values from $\{-1, 1\}^p$ with a distribution of the following
form:
\begin{equation} \label{eq:model_mrf}
\mathbb{P}_{\thetav^t}(\xv) = \frac{1}{Z(\thetav^t)} \exp
\biggl(\sum_{(u,v) \in E^t} \theta_{uv}^tx_ux_v \biggr),
\end{equation}
where $Z(\thetav^t)$ is the partition function, $\thetav^t \in
\mathbb{R}^{p \choose 2}$ is the parameter vector, and $G^t = (V, E^t)$
is an undirected graph representing conditional independence
assumptions among subsets of the $p$-dimensional random vector
$\Xv^t$. Recall that $V = \{1,\ldots,p\}$ is the node set and each
node corresponds with one component of the vector $\Xv^t$. In the
paper we are addressing the problem of graph structure estimation from
the observational data, which we now formally define: \textit{
given any time point $\tau \in [0, 1]$ estimate the graph structure
associated with $\mathbb{P}_{\thetav^t}$, given the observations
$\mathcal{D}_n$. } To obtain insight into the dynamics of changes in
the graph structure, one only needs to estimate graph structure for
multiple time-points, for example, for every $\tau \in \mathcal{T}_n$.

The graph structure $G^\tau$ is encoded by the locations of the
nonzero elements of the parameter vector $\thetav^\tau$, which we
refer to as the nonzero pattern of the parameter
$\thetav^\tau$. Components of the vector $\thetav^\tau$ are indexed by
distinct pairs of nodes and a component of the vector
$\theta_{uv}^\tau$ is nonzero if and only if the corresponding edge
$(u,v) \in E^\tau$. Throughout the rest of the paper we will focus on
estimation of the nonzero pattern of the vector $\thetav^\tau$ as a
way to estimate the graph structure. Let $\thetav_u^\tau$ be the
$(p-1)$-dimensional subvector of parameters
\[
\thetav_{ u}^{\tau}:=
\{\theta_{uv}^{\tau}\mid v \in V \setminus u\}
\]
associated with each node $u \in V$, and let $S^\tau(u)$ be the set of
edges adjacent to a node $u$ at a time point $\tau$:
\[
S^\tau(u):= \{ (u,v) \in V \times V \mid \theta_{uv}^\tau \neq 0\}.
\]
Observe that the graph structure $G^\tau$ can be recovered from the
local information on neighboring edges $S^\tau(u)$, for each node $u
\in V$, which can be obtained from the nonzero pattern of the
subvector $\thetav_u^\tau$ alone. The main focus of this section is on
obtaining node-wise estimators $\hat{\thetav}{}_u^\tau$ of the nonzero
pattern of the subvector $\thetav_u^\tau$, which are then used to
create estimates
\begin{equation}
\hat S^\tau(u):= \{ (u,v) \in V \times V \mid \hat \theta_{uv}^\tau \neq 0\},\qquad u \in V.
\end{equation}
Note that the estimated nonzero pattern might be asymmetric, for example,
$\hat \theta_{uv}^\tau = 0$, but $\hat \theta_{vu}^\tau \neq 0$. We
consider using the $\min$ and $\max$ operations to combine the
estimators $\hat \theta_{uv}^\tau$ and $\hat \theta_{vu}^\tau$. Let
$\tilde{\thetav}{}^\tau$ denote the combined estimator. The estimator
combined using the $\min$ operation has the following form:
\begin{equation} \label{eq:min_symmetrization}
\tilde \theta_{uv} =
\cases{
\hat \theta_{uv}, &\quad\mbox{if }$|\hat \theta_{uv}| < |\hat\theta_{vu}|$,\cr
\hat \theta_{vu}, &\quad\mbox{if }$|\hat \theta_{uv}| \geq|\hat\theta_{vu}|$,
}
\qquad
\mbox{``min\_symmetrization,''}
\end{equation}
which means that the edge $(u,v)$ is included in the graph estimate
only if it appears in both estimates $\hat S^\tau(u)$ and $\hat
S^\tau(v)$. Using the $\max$ operation, the combined estimator can be
expressed as
\begin{equation} \label{eq:max_symmetrization}
\tilde \theta_{uv} =
\cases{
\hat \theta_{uv}, &\quad\mbox{if }$|\hat \theta_{uv}| >|\hat\theta_{vu}|$,\cr
\hat \theta_{vu}, &\quad\mbox{if }$|\hat \theta_{uv}| \leq|\hat\theta_{vu}|$,
}
\qquad\mbox{``max\_symmetrization,''}
\end{equation}
and, as a result, the edge $(u,v)$ is included in the graph estimate if
it appears in at least one of the estimates $\hat S^\tau(u)$ or $\hat
S^\tau(v)$.

An estimator $\hat{\thetav}{}_u^\tau$ is obtained through the use of
pseudo-likelihood based on the conditional distribution of $X_u^\tau$
given the other of variables $\Xv_{\setminus u}^\tau = \{ X_v^\tau \mid v \in V \setminus u\}$. Although the use of pseudo-likelihood
fails in certain scenarios, for example, estimation of Exponential Random
Graphs [see \citet{vanDuijn09framework} for a recent study], the graph
structure of an Ising model can be recovered from an i.i.d.
sample using the pseudo-likelihood, as shown in
\citet{ravikumar09high}.  Under the model \eqref{eq:model_mrf}, the
conditional distribution of $X_u^{\tau}$ given the other variables
$\Xv_{\setminus u}^{\tau}$ takes the form
\begin{equation} \label{eq:cond_likelihood}
\mathbb{P}_{\thetav_u^{\tau}}(x_{u}^{\tau} |
\Xv_{\setminus u}^{\tau} = \xv_{\setminus u}^{\tau}) =
\frac{ \exp (x_u^{\tau} \langle\thetav_{ u}^{\tau},
\xv_{\setminus u}^{\tau} \rangle ) }
{ \exp (x_u^{\tau} \langle\thetav_{ u}^{\tau},
\xv_{\setminus u}^{\tau} \rangle ) +
\exp (-x_u^{\tau} \langle\thetav_{ u}^{\tau},
\xv_{\setminus u}^{\tau} \rangle ) },
\end{equation}
where $\langle \mathbf{a}, \mathbf{b} \rangle = \mathbf{a}'\mathbf{b}$
denotes the dot product. For simplicity, we will write
$\mathbb{P}_{\thetav_u^{\tau}}(x_{u}^{\tau} | \Xv_{\setminus
u}^{\tau} = \xv_{\setminus u}^{\tau})$ as
$\mathbb{P}_{\thetav_u^{\tau}}(x_{u}^{\tau} | \xv_{\setminus
u}^{\tau})$. Observe that the model given in
equation~\eqref{eq:cond_likelihood} can be viewed as expressing
$X_u^{\tau}$ as the response variable in the generalized
varying-coefficient models with $\Xv_{\setminus u}^{\tau}$ playing
the role of covariates. Under the model given in
equation~\eqref{eq:cond_likelihood}, the conditional log-likelihood, for
the node $u$ at the time point $t \in \mathcal{T}_n$, can be written in
the following form:
\begin{eqnarray} \label{eq:cond_loglikelihood}
\gamma(\thetav_{ u}; \xv^t)
& = & \log \mathbb{P}_{\thetav_u}(x_u^t |\xv_{\setminus u}^t)\nonumber \\[-8pt]\\[-8pt]
& = & x_u^t \langle \thetav_{ u}, \xv_{\setminus u}^t \rangle
- \log \bigl(\exp ( \langle \thetav_{ u}, \xv_{\setminus u}^t\rangle)
+\exp ( -\langle \thetav_{ u}, \xv_{\setminus u}^t\rangle )\bigr).\nonumber
\end{eqnarray}
The nonzero pattern of $\thetav_u^\tau$ can be estimated by
maximizing the conditional log-likelihood given in
equation~\eqref{eq:cond_loglikelihood}. What is left to show is how to combine
the information across different time points, which will depend on the
assumptions that are made on the unknown vector $\thetav^t$.

The primary focus is to develop methods applicable to data sets with
the total number of observations $n$ small compared to the
dimensionality $p = p_n$. Without assuming anything about $\thetav^t$,
the estimation problem is ill-posed, since there can be more
parameters than samples. A common way to deal with the estimation
problem is to assume that the graphs $\{G^t\}_{t \in \mathcal{T}_n}$ are
sparse, that is, the parameter vectors $\{\thetav^t\}_{t \in \mathcal{T}_n}$
have only few nonzero elements. In particular, we assume that each
node $u$ has a small number of neighbors, that is, there exists a number
$s \ll p$ such that it upper bounds the number of edges $|S^\tau(u)|$
for all $u \in V$ and $\tau \in \mathcal{T}_n$. In many real data sets
the sparsity assumption holds quite well. For example, in a genetic
network, rarely a regulator gene would control more than a handful of
regulatees under a specific condition~\citep{davidson}. Furthermore,
we will assume that the parameter vector $\thetav^t$ behaves ``nicely''
as a function of time. Intuitively, without any assumptions about the
parameter $\thetav^t$, it is impossible to aggregate information from
observations even close in time, because the underlying probability
distributions for observations from different time points might be
completely different. In the paper we will consider two ways of
constraining the parameter vector $\thetav^t$ as a function of time:
\begin{itemize}
\item \textit{Smooth changes in parameters.} We first consider that the
distribution generating the observation changes smoothly over the
time, that is, the parameter vector $\thetav^t$ is a smooth function of
time. Formally, we assume that there exists a constant $M > 0$ such
that it upper bounds the following quantities:
\[
\max_{u,v \in V \times V} \sup_{t \in [0,1]}
\bigg|\frac{\partial}{\partial t} \theta_{uv}^t\bigg| < M,
\qquad
\max_{u,v \in V \times V} \sup_{t \in [0,1]}
\bigg|\frac{\partial^2}{\partial t^2} \theta_{uv}^t\bigg| < M.
\]
Under this assumption, as we get more and more data (i.e., we collect
data in higher and higher temporal resolution within interval
$[0,1]$), parameters, and graph structures, corresponding to any two
adjacent time points will differ less and less.
\item \textit{Piecewise constant with abrupt structural changes in parameters.} Next, we consider that
there are a number of change points at which the distribution
generating samples changes abruptly. Formally, we assume that, for
each node $u$, there is a partition $\mathcal{B}_u = \{ 0 = B_{u, 0} <
B_{u, 1} < \cdots < B_{u, k_u} = 1 \}$ of the interval $[0, 1]$,
such that each element of $\thetav_u^t$ is constant on each segment
of the partition. At change points some of the elements of the
vector $\thetav_u^t$ may become zero, while some others may become
nonzero, which corresponds to a change in the graph structure. If
the number of change points is small, that is, the graph structure
changes infrequently, then there will be enough samples at a segment
of the partition to estimate the nonzero pattern of the vector
$\thetav^\tau$.
\end{itemize}
In the following two subsections we propose two estimation methods,
each suitable for one of the assumptions discussed above.

\subsection{Smooth changes in parameters} \label{sec:methods:smooth}

Under the assumption that the elements of $\thetav^t$ are smooth
functions of time, as described in the previous section, we use a
kernel smoothing approach to estimate the nonzero pattern of
$\thetav_u^\tau$ at the time point of interest $\tau \in [0,1]$, for
each node $u \in V$. These node-wise estimators are then combined
using either equation~\eqref{eq:min_symmetrization} or
equation~\eqref{eq:max_symmetrization} to obtain the estimator of the
nonzero pattern of $\thetav^\tau$. The estimator $\hat{\thetav}{}_u^\tau$ is defined as a minimizer of the following objective:
\begin{equation} \label{eq:opt_problem}
\hat{\thetav}{}_u^\tau:=
\min_{\thetav_u \in \mathbb{R}^{p-1}}
 \{
l  (\thetav_u; \mathcal{D}_n ) + \lambda_1 \norm{\thetav_u}_1
 \},
\end{equation}
where
\begin{equation} \label{eq:weighted_logloss}
l(\thetav_u; \mathcal{D}_n) = - \sum_{t \in \mathcal{T}_n} w_t^\tau
\gamma(\thetav_u; \xv^t)
\end{equation}
is a weighted log-likelihood, with weights defined as $ w_t^\tau =
\frac{ K_{h}  (t-\tau ) } {\sum_{t' \in \mathcal{T}_n} K_{h}
 (t'-\tau )}$ and $K_{h}(\cdot) = K(\cdot/h)$ is a
symmetric, nonnegative kernel function. We will refer to this approach
of obtaining an estimator as \verb|smooth|. The $\ell_1$ norm of the
parameter is used to regularize the solution and, as a result, the
estimated parameter has a lot of zeros. The number of the nonzero
elements of $\hat{\thetav}{}_u^\tau$ is controlled by the user-specified
regularization parameter $\lambda_1 \geq 0$. The bandwidth parameter
$h$ is also a user defined parameter that effectively controls the
number of observations around $\tau$ used to obtain $\hat{\thetav}{}_u^\tau$. In Section~\ref{sec:tuning} we discuss how to choose
the parameters $\lambda_1$ and $h$.

The optimization problem \eqref{eq:opt_problem} is the well-known
objective of the $\ell_1$ penalized logistic regression and there are
many ways of solving it, for example, the interior point method of
\citet{koh07interior}, the projected subgradient descent method of
\citet{Duchi08projected}, or the fast coordinate-wise descent method of
\citet{friedman08regularization}. From our limited experience, the
specialized first order methods work faster than the interior point
methods and we briefly describe the iterative coordinate-wise descent
method:
\begin{enumerate}[1.]
\item Set initial values: $\hat{\thetav}{}_u^{\tau,0} \leftarrow \mathbf{0}$.
\item For each $v \in V \setminus u$, set the current estimate $\hat
\theta_{uv}^{\tau,\mathit{iter}+1}$ as a solution to the
following optimization procedure:
\begin{equation} \label{eq:l1_iterative}
\min_{\theta \in \mathbb{R}}  \biggl\{
\sum_{t \in \mathcal{T}_n} \gamma ( \hat
\theta_{u,1}^{\tau,\mathit{iter}+1}, \ldots,
\hat \theta_{u,v-1}^{\tau,\mathit{iter}+1},
\theta,
\hat \theta_{u,v+1}^{\tau,\mathit{iter}}, \ldots,
\hat \theta_{u,p-1}^{\tau,\mathit{iter}}; \xv^t  )
+ \lambda_1 |\theta| \biggr\}.
\end{equation}

\item Repeat step 2 until convergence
\end{enumerate}
For  an efficient way of solving \eqref{eq:l1_iterative} refer to
\citet{friedman08regularization}. In our experiments, we find that the
neighborhood of each node can be estimated in a few seconds even when
the number of covariates is up to a thousand. A nice property of our
algorithm is that the overall estimation procedure decouples to a
collection of separate neighborhood estimation problems, which can be
trivially parallelized. If we treat the neighborhood estimation as an
atomic operation, the overall algorithm scales linearly as a product
of the number of covariates $p$ and the~number of time points $n$,
that is, $\mathcal{O}(pn)$. For instance, the Drosophila data set in the
application section contains 588 genes and 66 time points. The method
\verb|smooth| can estimate the neighborhood of one node, for all
points in a regularization plane, in less than 1.5 hours.\footnote{We
have used a server with dual core 2.6GHz processor and 2GB RAM.}

\subsection{Structural changes in parameters} \label{sec:methods:tv}

In this section we give the estimation procedure of the nonzero
pattern of $\{\thetav^t\}_{t \in \mathcal{T}_n}$ under the assumption
that the elements of $\thetav_u^t$ are a piecewise constant function,
with pieces defined by the partition $\mathcal{B}_u$. Again, the
estimation is performed node-wise and the estimators are combined
using either equation~\eqref{eq:min_symmetrization} or
equation~\eqref{eq:max_symmetrization}. As opposed to the kernel smoothing
estimator defined in equation~\eqref{eq:opt_problem}, which gives the
estimate at one time point $\tau$, the procedure described below
simultaneously estimates $\{ \hat{\thetav}{}_u^t \}_{t \in \mathcal{
T}_n}$. The estimators $\{ \hat{\thetav}{}_u^t \}_{t \in \mathcal{T}_n}$
are defined as a minimizer of the following convex optimization
objective:
\begin{equation} \label{eq:varying_coeff_tv}
\hspace*{25pt}\argmin_{\thetav_u^t \in \mathbb{R}^{p-1}, t \in \mathcal{T}_n }
\biggl\{ \sum_{t \in \mathcal{T}_n} \gamma(\thetav_u^t; \xv^t) +
\lambda_1 \sum_{t \in \mathcal{T}_n} \norm{\thetav_u^t}_1 +
\lambda_{\TV} \sum_{v \in V \setminus u}
\TV(\{\theta_{uv}^t\}_{t \in \mathcal{T}_n}) \biggr\},
\end{equation}
where $\TV(\{\theta_{uv}^t\}_{t \in \mathcal{T}_n}):= \sum_{i = 2}^n
|\theta_{uv}^{i/n} - \theta_{uv}^{(i-1)/n}|$ is the total variation
penalty.  We will refer to this approach of obtaining an estimator as
\verb|TV|. The penalty is structured as a combination of two terms. As
mentioned before, the $\ell_1$ norm of the parameters is used to
regularize the solution toward estimators with lots of zeros and the
regularization parameter $\lambda_1$ controls the number of nonzero
elements. The second term penalizes the difference between parameters
that are adjacent in time and, as a result, the estimated parameters
have infrequent changes across time. This composite penalty, known as
the ``fused'' Lasso penalty, was successfully applied in a slightly
different setting of signal denoising [e.g.,
\citet{Rinaldo08properties}] where it creates an estimate of the signal
that is piecewise constant.

The optimization problem given in equation~\eqref{eq:varying_coeff_tv} is
convex and can be solved using an off-the-shelf interior point solver
[e.g., the \texttt{CVX} package by \citet{cvx}]. However, for large
scale problems (i.e., both $p$ and $n$ are large), the interior point
method can be computationally expensive, and we do not know of any
specialized algorithm that can be used to
solve~\eqref{eq:varying_coeff_tv} efficiently. Therefore, we propose a
block-coordinate descent procedure which is much more efficient than
the existing off-the-shelf solvers for large scale problems. Observe
that the loss function can be decomposed as
$\mathcal{L}(\{\thetav_u^t\}_{t \in \mathcal{T}_n}) =
f_1(\{\thetav_u^t\}_{t \in \mathcal{T}_n}) +
\sum_{v \in V\setminus u} f_2(\{\theta_{uv}^t\}_{t \in \mathcal{T}_n})$
for a smooth differentiable convex function
$ f_1(\{\thetav_u^t\}_{t \in \mathcal{T}_n}) = \sum_{t \in \mathcal{T}_n}
\gamma(\thetav_u^t; \xv^t) $
and a convex function
$ f_2(\{\theta_{uv}^t\}_{t \in \mathcal{T}_n}) = \lambda_1 \sum_{t \in
\mathcal{T}_n} |\theta_{uv}^t| + \lambda_{\TV}
\TV(\{\theta_{uv}^t\}_{t \in \mathcal{T}_n}).  $
\citet{tseng01convergence} established that the
block-coordinate descent converges for loss functions with such
structure. Based on this observation, we propose the following
algorithm:
\begin{enumerate}
\item Set initial values: $\hat{\thetav}{}_{u}^{t,0} \leftarrow
\mathbf{0},   \forall t \in \mathcal{T}_n$.
\item For each $v \in V \setminus u$, set the current estimates $\{
\hat \theta_{uv}^{t,\mathit{iter}+1}\}_{t \in \mathcal{T}_n}$ as a solution to the
following optimization procedure:
\begin{eqnarray} \label{eq:tv_iterative}
&&\min_{\{\theta^t \in \mathbb{R}\}_{t \in \mathcal{T}_n} }
 \biggl\{
\sum_{t \in \mathcal{T}_n} \gamma (
\hat \theta_{u,1}^{t,\mathit{iter}+1}, \ldots, \hat \theta_{u,v-1}^{t,\mathit{iter}+1},
\theta^t,
\hat \theta_{u,v+1}^{t,\mathit{iter}}, \ldots, \hat \theta_{u,p-1}^{t,\mathit{iter}};
\xv^t  )\nonumber \\[-8pt]\\[-8pt]
&&\hspace*{136pt}{}+ \lambda_1 \sum_{t \in \mathcal{T}^n} |\theta^t|
+ \lambda_{\TV} \TV(\{\theta^t\}_{t \in \mathcal{T}_n})
 \biggr\}.\nonumber
\end{eqnarray}
\item Repeat step 2 until convergence.
\end{enumerate}

Using the proposed block-coordinate descent algorithm, we solve a
sequence of optimization problems each with only $n$ variables given
in equation~\eqref{eq:tv_iterative}, instead of solving one big
optimization problem with $n(n-1)$ variables given in
equation~\eqref{eq:varying_coeff_tv}. In our experiments, we find that the
optimization in~equation~\eqref{eq:varying_coeff_tv} can be estimated in an
hour when the number of covariates is up to a few hundred and when the
number of time points is also in the hundreds. Here, the bottleneck is the
number of time points. Observe that the dimensionality of the problem
in equation~\eqref{eq:tv_iterative} grows linearly with the number of time
points.  Again, the overall estimation procedure decouples to a
collection of smaller problems which can be trivially parallelized. If
we treat the optimization in~equation~\eqref{eq:varying_coeff_tv} as an
atomic operation, the overall algorithm scales linearly as a function
of the number of covariates $p$, that is, $\mathcal{O}(p)$. For instance, the
Senate data set in the application section contains 100 Senators and
542 time points. It took about a day to solve the optimization problem
in equation~\eqref{eq:varying_coeff_tv} for all points in the regularization
plane.

\subsection{Multiple observations}

In the discussion so far, it is assumed that at any time point in
$\mathcal{T}_n$ only one observation is available. There are situations
with multiple observations at each time point, for example, in a controlled
repeated microarray experiment two samples obtained at a certain time
point could be regarded as independent and identically distributed,
and we discuss below how to incorporate such observations into our
estimation procedures. Later, in Section~\ref{sec:simulation} we
empirically show how the estimation procedures benefit from additional
observations at each time point.

For the estimation procedure given in equation~\eqref{eq:opt_problem}, there
are no modifications needed to accommodate multiple observations at a
time point. Each additional sample will be assigned the same weight
through the kernel function $K_{h}(\cdot)$. On the other hand, we need
a small change in equation~\eqref{eq:varying_coeff_tv} to allow for
multiple observations. The estimators $\{ \hat{\thetav}{}_u^t \}_{t \in
\mathcal{T}_n}$ are defined as follows:
\begin{eqnarray}
&&\argmin_{\thetav_u^t \in \mathbb{R}^{p-1}, t \in \mathcal{T}_n }
\biggl\{ \sum_{t \in \mathcal{T}_n}
\sum_{\xv \in \mathcal{D}_n^t}
\gamma(\thetav_u^t; \xv)\nonumber
\\[-8pt]\\[-8pt]
&&\hspace*{35pt}\qquad{}+\lambda_1 \sum_{t \in \mathcal{T}_n} \norm{\thetav_u^t}_1 +
\lambda_{\TV} \sum_{v \in V \setminus u}
\TV(\{\theta_{uv}^t\}_{t \in \mathcal{T}_n}) \biggr\},\nonumber
\end{eqnarray}
where the set $\mathcal{D}_n^t$ denotes elements from the sample
$\mathcal{D}_n$ observed at a time point $t$.

\subsection{Choosing tuning parameters} \label{sec:tuning}

Estimation procedures discussed in Sections~\ref{sec:methods:smooth}
and~\ref{sec:methods:tv}, \verb|smooth| and \verb|TV| respectively,
require a choice of tuning parameters. These tuning parameters control
sparsity of estimated graphs and the way the graph structure changes
over time. The tuning parameter $\lambda_1$, for both \verb|smooth|
and \verb|TV|, controls the sparsity of the graph structure. Large
values of the parameter $\lambda_1$ result in estimates with lots of
zeros, corresponding to sparse graphs, while small values result in
dense models. Dense models will have a higher pseudo-likelihood score,
but will also have more degrees of freedom. A good choice of the
tuning parameters is essential in obtaining a good estimator that does
not overfit the data, and balances between the pseudo-likelihood and
the degrees of freedom. The bandwidth parameter $h$ and the penalty
parameter $\lambda_{\TV}$ control how similar are estimated networks
that are close in time. Intuitively, the bandwidth parameter controls
the size of a window around time point $\tau$ from which observations
are used to estimate the graph $G^\tau$. Small values of the bandwidth
result in estimates that change often with time, while large values
produce estimates that are almost time invariant. The penalty
parameter $\lambda_{\TV}$ biases the estimates $\{\hat{\thetav}{}_u^t\}_{t \in \mathcal{T}_n}$ that are close in time to have
similar values; large values of the penalty result in graphs whose
structure changes slowly, while small values allow for more changes in
estimates.

First, we discuss how to choose the penalty parameters $\lambda_1$ and
$\lambda_{\TV}$ for the method \verb|TV|. Observe that
$\gamma(\thetav_u^t; \xv^t)$ represents a logistic regression loss
function when regressing a node $u$ onto the other nodes $V \setminus
u$. Hence, problems defined in equation~\eqref{eq:opt_problem} and
equation~\eqref{eq:varying_coeff_tv} can be regarded as \textit{supervised}
classification problems, for which a number of techniques can be used
to select the tuning parameters, for example, cross-validation or held-out
data sets can be used when enough data is available, otherwise, the BIC
score can be employed. In this paper we focus on the BIC score
defined for $\{\thetav_u^t\}_{t \in \mathcal{T}_n}$ as
\begin{equation} \label{eq:BIC}
\operatorname{BIC}(\{\thetav_u^t\}_{t \in \mathcal{T}_n}):=
\sum_{t \in \mathcal{T}_n} \gamma(\thetav_u^t; \xv^t) -
\frac{\log n}{2} \operatorname{Dim}(\{\thetav_u^t\}_{t \in \mathcal{T}_n}),
\end{equation}
where $\operatorname{Dim}(\cdot)$ denotes the degrees of freedom of the
estimated model. Similar to \citep{tibshirani05sparsity}, we adopt the
following approximation to the degrees of freedom:
\begin{eqnarray} \label{eq:BIC:dim}
\operatorname{Dim}(\{\thetav_u^t\}_{t \in \mathcal{T}_n})
&=&\sum_{t \in \mathcal{T}_n}
\sum_{v \in V \setminus u} \ind  [
\sign( \theta^{t}_{uv}) \neq \sign (\theta^{t-1}_{uv})  ]\nonumber
\\[-8pt]\\[-8pt]
&&\hspace*{41pt}{}\times \ind  [\sign(\theta^t_{uv}) \neq 0  ],\nonumber
\end{eqnarray}
which counts the number of blocks on which the parameters are constant
and not equal to zero. In practice, we average the BIC scores from all
nodes and choose models according to the average.

Next, we address the way to choose the bandwidth $h$ and the penalty
parameter $\lambda_1$ for the method \verb|smooth|. As mentioned
earlier, the tuning of bandwidth parameter $h$ should trade off the
smoothness of the network changes and the coverage of samples used to
estimate the network. Using a wider bandwidth parameter provides more
samples to estimate the network, but this risks missing sharper
changes in the network; using a narrower bandwidth parameter makes the
estimate more sensitive to sharper changes, but this also makes the
estimate subject to larger variance due to the reduced effective
sample size.  In this paper we adopt a heuristic for tuning the
inital scale of the bandwidth parameter: we set it to be the median of
the distance between pairs of time points. That is, we first form a
matrix $(d_{ij})$ with its entries $d_{ij}:=(t_i-t_j)^2$
($t_i,t_j\in\mathcal{T}_{n}$). Then the scale of the bandwidth
parameter is set to the median of the entries in $(d_{ij})$. In our
later simulation experiments, we find that this heuristic provides a
good initial guess for $h$, and it is quite close to the value
obtained via exhaustive grid search. For the method \verb|smooth|, the
BIC score for $\{\thetav_u^t\}_{t \in \mathcal{T}_n}$ is defined as
\begin{equation} \label{eq:BIC_smooth}
\operatorname{BIC}(\{\thetav_u^t\}_{t \in \mathcal{T}_n}):=
\sum_{\tau \in \mathcal{T}_n} \sum_{t \in \mathcal{T}_n}
w_t^\tau \gamma(\thetav_u^\tau; \xv^t) -
\frac{\log n}{2} \operatorname{Dim}(\{\thetav_u^t\}_{t \in \mathcal{T}_n}),
\end{equation}
where $\operatorname{Dim}(\cdot)$ is defined in equation~\eqref{eq:BIC:dim}.

\section{Simulation studies} \label{sec:simulation}

We have conducted a small empirical study of the performance of
methods \verb|smooth| and \verb|TV|. Our idea was to choose parameter
vectors $\{\thetav^t\}_{t \in \mathcal{T}_n}$, generate data according
to the model in equation~\eqref{eq:model_mrf} using Gibbs sampling, and try
to recover the nonzero pattern of $\thetav^t$ for each $t \in \mathcal{
T}_n$. Parameters $\{\thetav^t\}_{t \in \mathcal{T}_n}$ are considered
to be evaluations of the function $\thetav^t$ at $\mathcal{T}_n$ and we
study two scenarios, as discussed in Section~\ref{sec:methods}:
$\thetav^t$ is a smooth function, $\thetav^t$ is a piecewise constant
function. In addition to the methods \verb|smooth| and \verb|TV|, we
will use the method of \citet{ravikumar09high} to estimate a
time-invariant graph structure, which we refer to as
\verb|static|. All of the three methods estimate the graph based on
node-wise neighborhood estimation, which, as discussed in
Section~\ref{sec:methods}, may produce asymmetric
estimates. Solutions combined with the min operation in
equation~\eqref{eq:min_symmetrization} are denoted as ${*}{*}{*}{*}$\verb|.MIN|, while
those combined with the max operation in
equation~\eqref{eq:max_symmetrization} are denoted as ${*}{*}{*}{*}$\verb|.MAX|.

We took the number of nodes $p = 20$, the maximum node degree $s = 4$,
the number of edges $e = 25,$ and the sample size $n = 500$. The
parameter vectors $\{\thetav^t\}_{t\in\mathcal{T}_n}$ and observation
sequences are generated as follows:
\begin{enumerate}
\item Generate a random graph $\tilde G^0$ with $20$ nodes and 15
edges: edges are added, one at a time, between random pairs of
nodes that have the node degree less than 4.  Next, randomly add 10
edges and remove 10 edges from $\tilde G^0$, taking care that the
maximum node degree is still $4$, to obtain $\tilde G^1$. Repeat the
process of adding and removing edges from $\tilde G^1$ to obtain
$\tilde G^2, \ldots, \tilde G^5$. We refer to these 6 graphs as the
anchor graphs. We will randomly generate the prototype parameter
vectors $\tilde{\thetav}{}^0, \ldots, \tilde{\thetav}{}^5$, corresponding
to the anchor graphs, and then interpolate between them to obtain
the parameters $\{\thetav^t\}_{t\in\mathcal{T}_n}$.
\item Generate a prototype parameter vector $\tilde{\thetav}{}^i$ for
each anchor graph $\tilde G^i$, $i \in \{0, \ldots, 5\}$, by
sampling nonzero elements of the vector independently from $\mathsf{Unif}([0.5, 1])$. Then generate $\{\thetav^t\}_{t \in \mathcal{T}_n}$
according to one of the following two cases:
\begin{itemize}
\item Smooth function: The parameters $\{ \thetav^t\}_{t \in
((i-1)/5, i/5] \cap \mathcal{T}_n}$ are obtained by linearly
interpolating $100$ points between $\tilde{\thetav}{}^{i-1}$ and
$\tilde{\thetav}{}^{i}$, $i \in \{1, \ldots, 5\}$.
\item Piecewise constant function: The parameters $\{ \thetav^t\}_{t
\in ((i-1)/5, i/5] \cap \mathcal{T}_n}$ are set to be equal to
$(\tilde{\thetav}{}^{i-1} + \tilde{\thetav}{}^{i})/2$, $i \in \{1,
\ldots, 5\}$.
\end{itemize}
Observe that after interpolating between the prototype parameters, a
graph corresponding to $\thetav^t$ has $25$ edges and the maximum
node degree is $4$.
\item Generate 10 independent samples at each $t \in \mathcal{T}_n$ according to
$\mathbb{P}_{\thetav^t}$, given in equation~\eqref{eq:model_mrf}, using
Gibbs sampling.
\end{enumerate}
We estimate $\hat G^t$ for each $t \in \mathcal{T}_n$ with our
\verb|smooth| and \verb|TV| methods, using $k \in \{1, \ldots, 10\}$
samples at each time point. The results are expressed in terms of the
precision~$(\mathsf{Pre})$ and the recall~$(\mathsf{Rec})$ and $F1$ score,
which is the harmonic mean of precision and recall, that is, $F1:= 2 *
\mathsf{Pre} * \mathsf{Rec} / (\mathsf{Pre} + \mathsf{Rec})$. Let $\hat E^t$
denote the estimated edge set of $\hat G^t$, then the precision is
calculated as $\mathsf{Pre}:= 1/n \sum_{t \in \mathcal{T}_n} |\hat E^t \cap
E^t|/|\hat E^t|$ and the recall as $\mathsf{Rec}:= 1/n \sum_{t \in \mathcal{
T}_n} |\hat E^t \cap E^t|/|E^t|$. Furthermore, we report results
averaged over 20 independent runs.

\begin{figure}[b]

\includegraphics{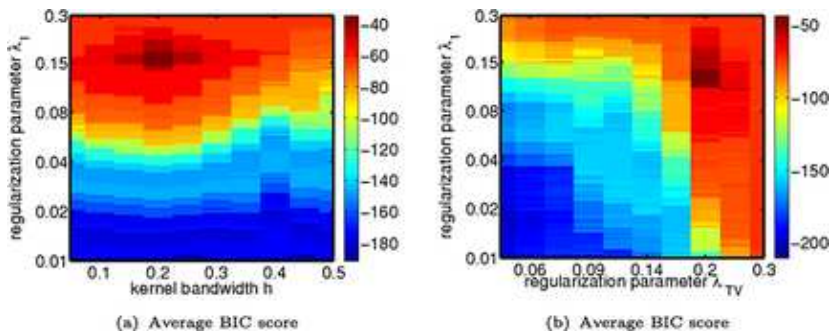}

\caption{Plot of the $\operatorname{BIC}_\mathrm{avg}$ score over the
regularization plane. The parameter vector $\theta^t$ is a smooth
function of time and at each time point there is one
observation.
\textup{(a)} 
The graph structure
recovered using the method $\operatorname{smooth}$.
\textup{(b)} 
Recovered using the method $\operatorname{TV}$.}\label{fig:bic_plot}
\end{figure}

The tuning parameters $h$ and $\lambda_1$ for \verb|smooth|, and
$\lambda_1$ and $\lambda_{\TV}$ for \verb|TV|, are chosen by maximizing
the average BIC score,
\[
\operatorname{BIC}_\mathrm{avg}:= 1/p \sum_{u \in V} \operatorname{BIC}(\{\thetav_u^t\}_{t \in \mathcal{T}_n}),
\]
over a grid of parameters. The bandwidth parameter $h$ is searched
over $\{ 0.05, 0.1,\ldots, 0.45, 0.5 \}$ and the penalty parameter
$\lambda_{\TV}$ over 10 points, equidistant on the log-scale, from the
interval $[0.05, 0.3]$. The penalty parameter is searched over 100
points, equidistant on the log-scale, from the interval $[0.01, 0.3]$
for both \verb|smooth| and \verb|TV|. The same range is used to select
the penalty parameter $\lambda$ for the method \verb|static| that
estimates a time-invariant network. In our experiments, we use the
Epanechnikov kernel $K(z) = 3/4 * (1-z^2)\ind\{|z| \leq 1\}$ and we
remind our reader that $K_h(\cdot) = K(\cdot/h)$. For illustrative
purposes, in Figure~\ref{fig:bic_plot} we plot the $\operatorname{BIC}_\mathrm{avg}$ score over the grid of tuning parameters.

\begin{figure}

\includegraphics{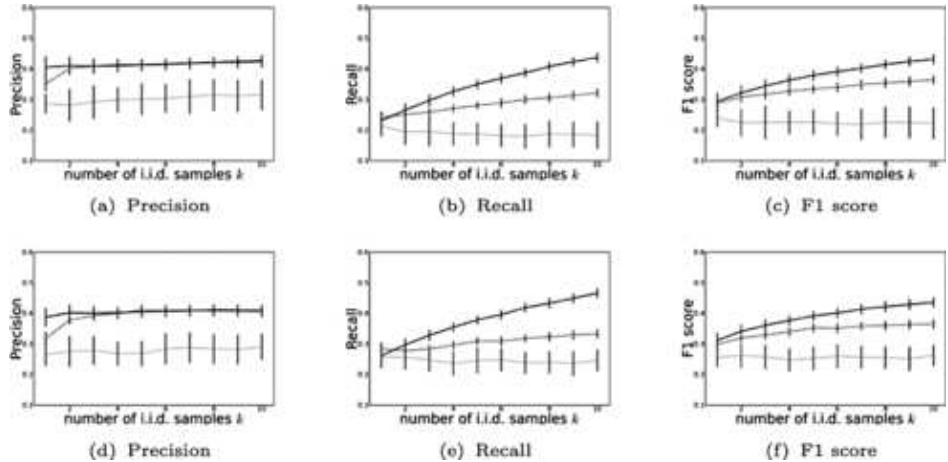}

\caption{Results of estimation when the underlying parameter
$\{\thetav^t\}_{t \in \mathcal{T}_n}$ changes smoothly with time. The
upper row consists of results when the graph is estimated
combining the neighborhoods using the $\min$ operation, while the
lower row consists of results when the $\max$ operation is used to
combine neighborhoods. Precision, recall, and F1 score are plotted
as the number of i.i.d. samples $k$ at each time point
increases from 1 to 10. The solid, dashed, and dotted lines denote
results for $\operatorname{smooth}$, $\operatorname{TV}$, and $\operatorname{static}$,
respectively.}
\label{fig:smooth_many}
\end{figure}

First, we discuss the estimation results when the underlying parameter
vector changes smoothly. See Figure~\ref{fig:smooth_many} for results.
It can be seen that as the number of the i.i.d. observations at
each time point increases, the performance of both methods
\verb|smooth| and \verb|TV| increases. On the other hand, the
performance of the method \verb|static| does not benefit from
additional i.i.d. observations. This observation should not be
surprising as the time-varying network models better fit the data
generating process. When the underlying parameter vector $\thetav^t$
is a smooth function of time, we expect that the method \verb|smooth|
would have a faster convergence and better performance, which can be
seen in Figure~\ref{fig:smooth_many}. There are some
differences between the estimates obtained through \verb|MIN| and
\verb|MAX| symmetrization. In our limited numerical experience, we
have seen that \verb|MAX| symmetrization outperforms \verb|MIN|
symmetrization. \verb|MIN| symmetrization is more conservative in
including edges to the graph and seems to be more susceptible to
noise.

\begin{figure}

\includegraphics{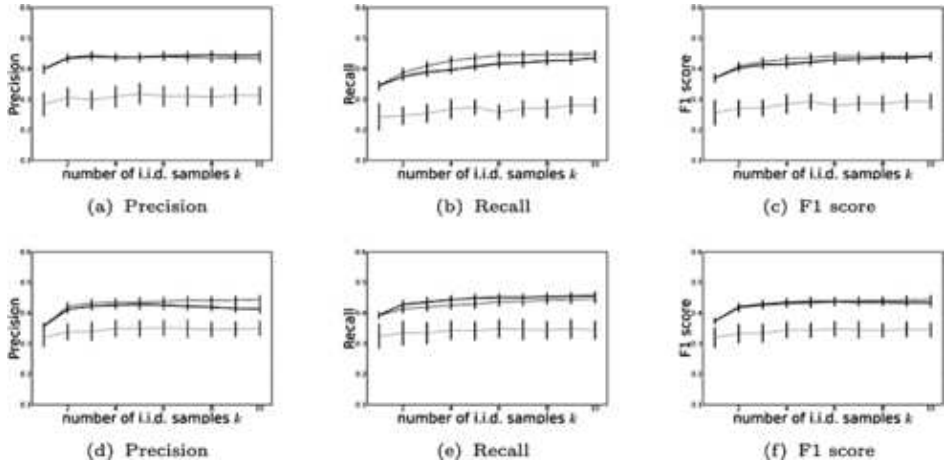}

\caption{Results of estimation when the underlying parameter
$\{\thetav^t\}_{t \in \mathcal{T}_n}$ is a piecewise constant
function of time. The upper row
consists of results when the graph is estimated combining the
neighborhoods using the $\min$ operation, while the lower row
consists of results when the $\max$ operation is used to combine
neighborhoods. Precision, recall, and F1 score are plotted as the
number of i.i.d. samples $k$ at each time point increases
from 1 to 10. The solid, dashed, and dotted lines denote results
for $\operatorname{smooth}$, $\operatorname{TV}$, and $\operatorname{static}$, respectively.}
\label{fig:jumps_many}
\end{figure}

Next, we discuss the estimation results when then the underlying
parameter vector is a piecewise constant function. See
Figure~\ref{fig:jumps_many} for results. Again, both performance of
the method \verb|smooth| and of the method \verb|TV| improve as there
are more independent samples at different time points, as opposed to
the method \verb|static|. It is worth noting that the empirical
performance of \verb|smooth| and \verb|TV| is very similar in the
setting when $\thetav^t$ is a piecewise constant function of time,
with the method \verb|TV| performing marginally better. This may be a
consequence of the way we present results, averaged over all time
points in $\mathcal{T}_n$. A closer inspection of the estimated graphs
shows that the method \verb|smooth| poorly estimates graph structure
close to the time point at which the parameter vector changes abruptly
(results not shown).\looseness=1

We have decided to perform simulation studies on Erd\"{o}s--R\'{e}nyi
graphs, while real-world graphs are likely to have different
properties, such as a scale-free network with a long tail in its
degree distribution. From a theoretical perspective (see
Section~\ref{sec:theory}), our method can still recover the true
structure of these networks regardless of the degree distribution,
although for a more complicated model, we may need more samples in
order to achieve this.  \citet{peng08partial} proposed a joint sparse
regression model, which performs better than the neighborhood
selection method when estimating networks with hubs (nodes with very
high degree) and scale-free networks. For such networks, we can
extend their model to our time-varying setting, and potentially make
more efficient use of the samples, however, we do not pursue this
direction here.

\section{Applications to real data} \label{sec:experiments}

In this section we present the analysis of two real data sets using
the algorithms presented in Section~\ref{sec:methods}. First, we
present the analysis of the senate data consisting of Senators' votes
on bills during the 109th Congress. The second data set consists of
expression levels of more than 4000 genes from the life cycle of
\textit{Drosophila melanogaster}.

\subsection{Senate voting records data}

The US senate data consists of voting records from 109th congress
(2005--2006).\footnote{The data can be obtain from the US Senate web
page \url{http://www.senate.gov}.}  There are 100 senators whose
votes were recorded on the 542 bills.  Each senator corresponds to a
variable, while the votes are samples recorded as $-$1 for no and 1 for
yes. This data set was analyzed in \citet{banerjee07model}, where a
static network was estimated. Here, we analyze this data set in a time-varying framework in order to discover how the relationship between
senators changes over time.

This data set has many missing values, corresponding to votes that
were not cast. We follow the approach of \citet{banerjee07model} and
fill those missing values with ($-$1). Bills were mapped onto the
$[0,1]$ interval, with 0 representing Jan 1st, 2005 and 1 representing
Dec 31st, 2006. We use the Epanechnikov kernel for the method
\verb|smooth|. The tuning parameters are chosen optimizing the average
BIC score over the same range as used for the simulations in
Section~\ref{sec:simulation}. For the method \verb|smooth|, the
bandwidth parameter was selected as $h = 0.174$ and the penalty
parameter $\lambda_1=0.195$, while penalty parameters $\lambda_1 =
0.24$ and $\lambda_{\TV}=0.28$ were selected for the method
\verb|TV|. In the figures in this section, we use pink square nodes
to represent republican Senators and blue circle nodes to represent
democrat Senators.

\begin{figure}

\includegraphics{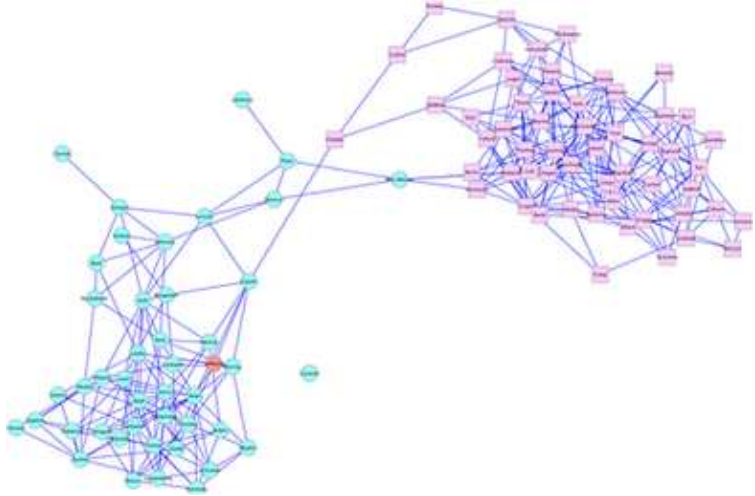}

\caption{109th Congress, Connections between Senators in April
2005. Democrats are represented with blue circles, Republicans
with pink squares, and the red circle represents independent
Senator Jeffords.}
\label{fig:complete_network}
\end{figure}

A first question is whether the learned network reflects the political
division between Republicans and Democrats. Indeed, at any time point
$t$, the estimated network contains few clusters of nodes. These
clusters consist of either Republicans or Democrats connected to each
others; see Figure~\ref{fig:complete_network}. Furthermore, there are
very few links connecting different clusters.  We observe that most
Senators vote similarly to other members of their party. Links
connecting different clusters usually go through senators that are
members of one party, but have views more similar to the other party,
for example, Senator Ben Nelson or Senator Chafee. Note that we do not
necessarily need to estimate a time evolving network to discover this
pattern of political division, as they can also be observed from a
time-invariant network, for example, see \citet{banerjee07model}.

\begin{figure}[b]

\includegraphics{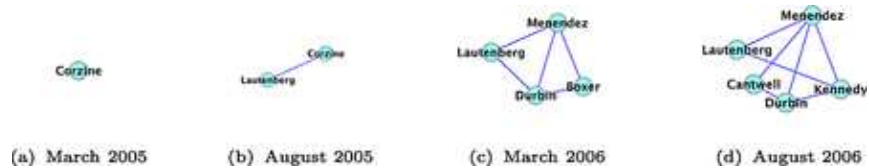}

\caption{Direct neighbors of the node that represent Senator
Corzine and Senator Menendez at four different time
points. Senator Corzine stepped down at the end of the 1st Session
and his place was taken by Senator Menendez, which is reflected in
the graph structure.}
\label{fig:Corzine}
\end{figure}

Therefore, what is more interesting is whether there is any time
evolving pattern.  To show this, we examine neighborhoods of Senators
Jon Corzine and Bob Menendez. Senator Corzine stepped down from the
Senate at the end of the 1st Session in the 109th Congress to become
the Governor of New Jersey. His place in the Senate was filled by
Senator Menendez.  This dynamic change of interactions can be well
captured by the time-varying
network~(Figure~\ref{fig:Corzine}). Interestingly, we can see that
Senator Lautenberg who used to interact with Senator Corzine switches to
Senator Menendez in response to this event.

\begin{figure}

\includegraphics{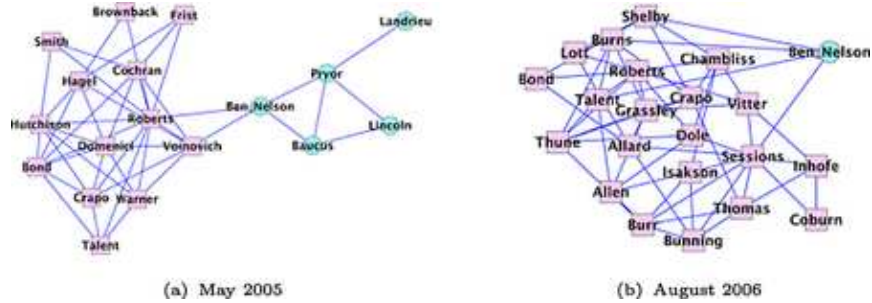}

\caption{Neighbors of Senator Ben Nelson (distance two or lower) at
the beginning of the 109th Congress and at the end of the 109th
Congress. Democrats are represented with blue circles, Republicans
with pink squares. The estimated neighborhood in August 2006
consists only of Republicans, which may be due to the type of
bills passed around that time on which Senator Ben Nelson had
similar views as other Republicans.}
\label{fig:Nelson}
\end{figure}

Another interesting question is whether we can discover senators with
swaying political stance based on time evolving networks. We discover
that Senator Ben Nelson and Lincoln Chafee fall into this
category. Although Senator Ben Nelson is a Democrat from Nebraska, he
is considered  to be one of the most conservative Democrats in the
Senate. Figure~\ref{fig:Nelson} presents neighbors at distance two or
less of Senator Ben Nelson at two time points, one during the 1st
Session and one during the 2nd Session. As a conservative Democrat, he
is connected to both Democrats and Republicans since he shares views
with both parties. This observation is supported by
Figure~\ref{fig:Nelson}(a) which presents his neighbors during the 1st
Session. It is also interesting to note that during the second
session, his views drifted more toward the Republicans
[Figure~\ref{fig:Nelson}(b)].  For instance, he voted against abortion
and withdrawal of most combat troops from Iraq, which are both
Republican views.

\begin{figure}[b]

\includegraphics{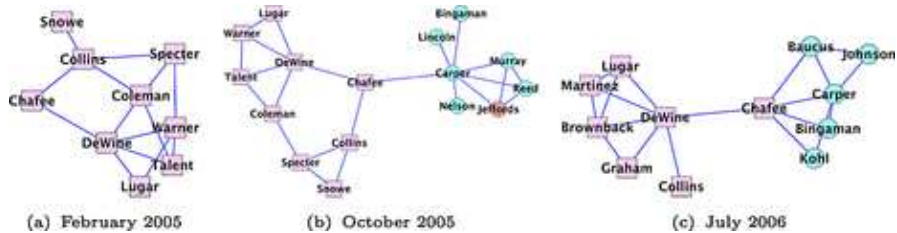}

\caption{Neighbors of Senator Chafee (distance two or lower) at
different time points during the 109th Congress. Democrats are
represented with blue circles, Republicans with pink squares, and
the red circle represents independent Senator Jeffords.}
\label{fig:Chafee}
\end{figure}
In contrast, although Senator Lincoln Chafee is a Republican, his
political view grew increasingly Democratic. Figure~\ref{fig:Chafee}
presents neighbors of Senator Chafee at three time points during the
109th Congress. We observe that his neighborhood includes an
increasing amount of Democrats as time progresses during the 109th
Congress. Actually, Senator Chafee later left the Republican Party and
became an independent in 2007. Also, his view on abortion, gay rights,
and environmental policies are strongly aligned with those of
Democrats, which is also consistently reflected in the estimated
network. We emphasize that these patterns about Senator Nelson and
Chafee could not be observed in a static network.

\subsection{Gene regulatory networks of Drosophila melanogaster}

In this section we used the kernel reweighting approach to reverse
engineer the gene regulatory networks of \textit{Drosophila melanogaster}
from a time series of gene expression data measured during its full
life cycle.  Over the developmental course of \textit{Drosophila
melanogaster}, there exist multiple underlying ``themes'' that
determine the functionalities of each gene and their relationships to
each other, and such themes are dynamical and stochastic. As a result,
the gene regulatory networks at each time point are context-dependent
and can undergo systematic rewiring, rather than being invariant over
time. In a seminal study by~\citet{LusBabYuEtal2004}, it was shown that
the ``active regulatory paths'' in the gene regulatory networks of
\textit{Saccharomyces cerevisiae} exhibit topological changes and hub
transience during a temporal cellular process, or in response to
diverse stimuli.  We expect similar properties can also be observed
for the gene regulatory networks of \textit{Drosophila melanogaster}.

We used microarray gene expression measurements
from~\citet{Arbeitmanetal2002} as our input data.  In such an
experiment, the expression levels of 4028 genes are simultaneously
measured at various developmental stages. Particularly, 66 time points
are chosen during the full developmental cycle of \textit{Drosophila
melanogaster}, spanning across four different stages, \textit{that is},
embryonic (1--30 time point), larval (31--40 time point), pupal
(41--58 time points), and adult stages (59--66 time points).  In this
study we focused on 588 genes that are known to be related to
the developmental process based on their gene ontologies.

\begin{figure}[b]

\includegraphics{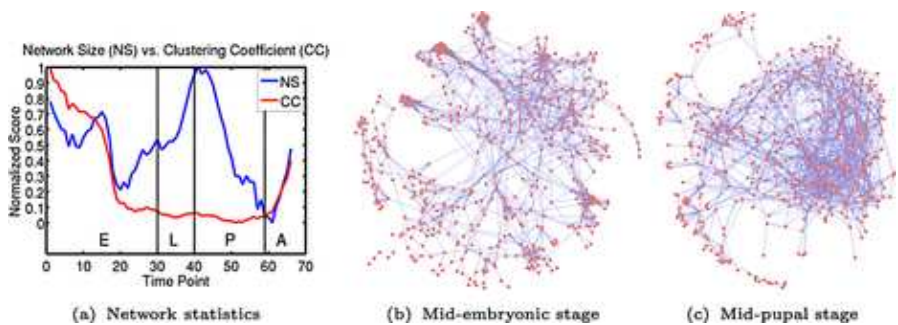}

\caption{Characteristic of the dynamic networks estimated for
the genes related to the developmental process. \textup{(a)} Plot of two
network statistics as functions of the development time
line. Network size ranges between 1712 and 2061 over time,
while local clustering coefficient ranges between 0.23 and
0.53 over time; To focus on relative activity over time,
both statistics are normalized to the range between 0 and
1. \textup{(b)} and \textup{(c)} are the visualization of two examples of networks from
different time points. We can see that network size can
evolve in a very different way from the local clustering
coefficient.}
\label{fig:overall}
\end{figure}

Usually, the samples prepared for microarray experiments are a mixture
of tissues with possibly different expression levels.  This means that
microarray experiments only provide rough estimates of the average
expression levels of the mixture.  Other sources of noise can also be
introduced into the microarray measurements during, for instance, the
stage of hybridization and digitization. Therefore, microarray
measurements are far from the exact values of the expression levels,
and it will be more robust if we only consider the binary state of the
gene expression: either being up-regulated or down-regulated.  For
this reason, we binarize the gene expression levels into $\{-1,1\}$
($-$1 for down-regulated and 1 for up-regulated).  We learned a sequence of
binary MRFs from these time series.

\begin{figure}[b]

\includegraphics{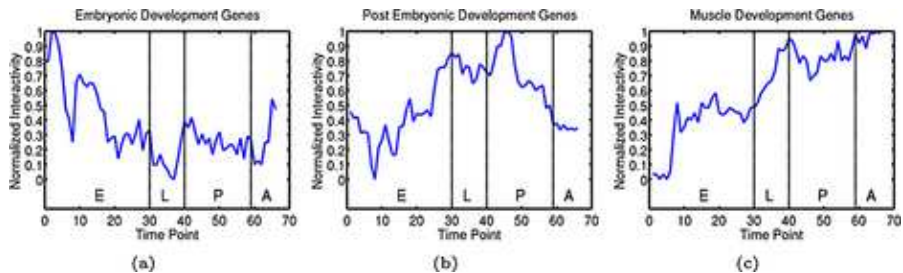}

\caption{Interactivity of 3 groups of genes related to \textup{(a)}
embryonic development (ranging between 169 and 241), \textup{(b)}
post-embryonic development (ranging between 120 and 210), and
\textup{(c)} muscle development (ranging between 29 and 89). To focus
on the relative activity over time, we normalize the score
to $[0,1]$. The higher the interactivity, the more active
the group of genes. The interactivities of these three
groups are very consistent with their functional
annotations.}
\label{fig:act}
\end{figure}

First, we study the global pattern of the time evolving regulatory networks.
In Figure~\ref{fig:overall}(a) we plotted two different statistics
of the reversed engineered gene regulatory networks as a function of
the developmental time point (1--66). The first statistic is the
network size as measured by the number of edges; and the second is the
average local clustering coefficient as defined
by~\citet{WatStro1998}. For comparison, we normalized both statistics
to the range between $[0,1]$. It can be seen that the network size and
its local clustering coefficient follow very different trajectories
during the developmental cycle. The network size exhibits a wave
structure featuring two peaks at mid-embryonic stage and the beginning
of the pupal stage. A similar pattern of gene activity has also been
observed by~\citet{Arbeitmanetal2002}. In contrast, the clustering
coefficients of the dynamic networks drop sharply after the
mid-embryonic stage, and they stay low until the start of the adult
stage. One explanation is that at the beginning of the development
process, genes have a more fixed and localized function, and they
mainly interact with other genes with similar functions. However,
after mid-embryonic stage, genes become more versatile and involved in
more diverse roles to serve the need of rapid development; as the
organism turns into an adult, its growth slows down and each gene is
restored to its more specialized role. To illustrate how the network
properties change over time, we visualized two networks from
mid-embryonic stage (time point 15) and mid-pupal stage (time point
45) using the spring layout algorithm in Figure~\ref{fig:overall}(b)
and~(c) respectively. Although the size of the two
networks are comparable, tight local clusters of interacting genes are
more visible during mid-embryonic stage than mid-pupal stage, which is
consistent with the evolution local clustering coefficient in
Figure~\ref{fig:overall}(a).

To judge whether the learned networks make sense biologically, we zoom
into three groups of genes functionally related to different stages of
the development process. In particular, the first group (30 genes) is
related to embryonic development based on their functional ontologies;
the second group (27 genes) is related to post-embryonic development;
and the third group (25 genes) is related to muscle development. For
each group, we use the number of within group connections plus all its
outgoing connections to describe the activitiy of each group of genes
(for short, we call it interactivity).  In Figure~\ref{fig:act} we
plotted the time courses of interactivity for the three groups
respectively. For comparison, we normalize all scores to the range of
$[0,1]$. We see that the time courses have a nice correspondence with
their supposed roles. For instance, embryonic development genes have
the highest interactivity during embryonic stage, and post-embryonic
genes increase their interactivity during the larval and pupal stages. The
muscle development genes are less specific to certain developmental
stages, since they are needed across the developmental cycle. However,
we see its increased activity when the organism approaches its adult
stage where muscle development becomes increasingly important.

\begin{figure}

\includegraphics{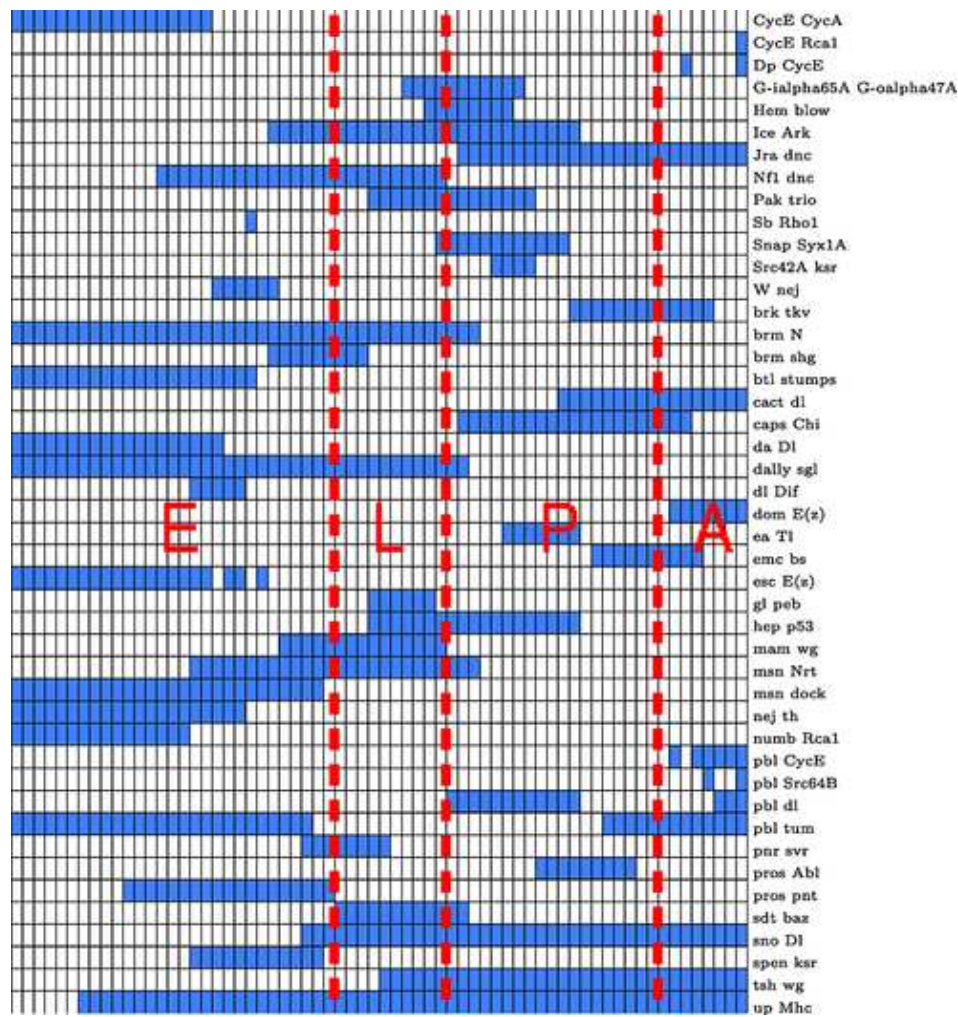}

\caption{Timeline of 45 known gene interactions. Each cell in
the plot corresponds to one gene pair of gene interaction at
one specific time point. The cells in each row are ordered
according to their time point, ranging from embryonic stage
\textup{(E)} to larval stage \textup{(L)}, to pupal stage \textup{(P)}, and to adult
stage \textup{(A)}. Cells colored blue indicate the corresponding
interaction listed in the right column is present in the
estimated network; blank color indicates the interaction is
absent. }
\label{fig:known_interaction}
\end{figure}

The estimated networks also recover many known interactions between
genes. In recovering these known interactions, the dynamic networks
also provide additional information as to when interactions occur
during development.  In Figure~\ref{fig:known_interaction} we listed
these recovered known interactions and the precise time when they
occur. This also provides a way to check whether the learned networks
are biologically plausible given the prior knowledge of the actual
occurrence of gene interactions. For instance, the interaction between
genes msn and dock is related to the regulation of embryonic cell
shape, correct targeting of photoreceptor axons.  This is very
consistent with the timeline provided by the dynamic networks. A
second example is the interaction between genes sno and Dl which is
related to the development of compound eyes of \textit{Drosophila}.  A
third example is between genes caps and Chi which are related to wing
development during pupal stage.  What is most interesting is that the
dynamic networks provide timelines for many other gene interactions
that have not yet been verified experimentally. This information will
be a useful guide for future experiments.

\begin{figure}

\includegraphics{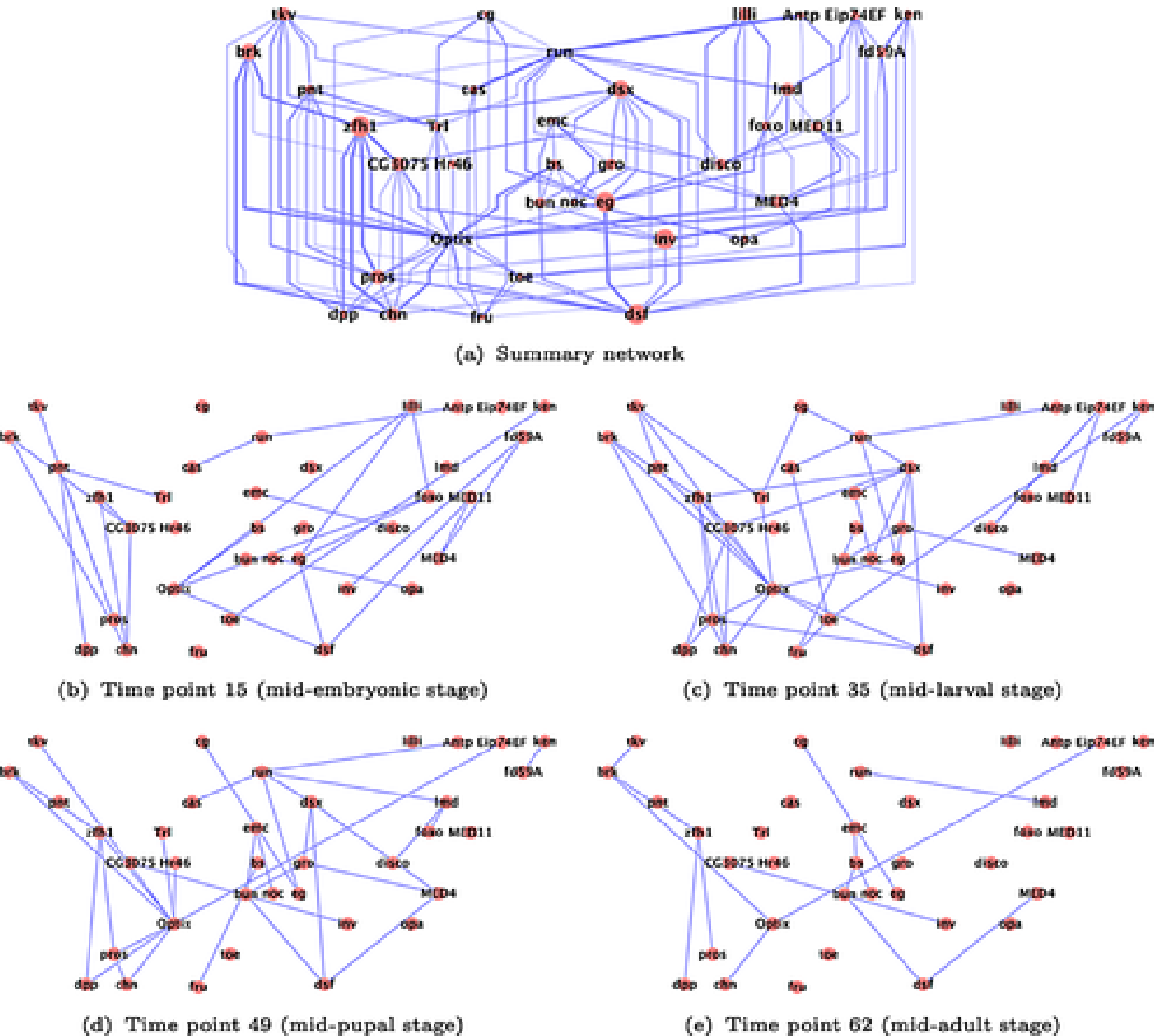}

\caption{The largest transcriptional factors (TF) cascade
involving 36 transcriptional factors. \textup{(a)}~The summary
network is obtained by summing the networks from all time
points. Each node in the network represents a
transcriptional factor, and each edge represents an
interaction between them.
On different stages of the development, the networks are
different, \textup{(b)}, \textup{(c)}, \textup{(d)}, \textup{(e)} shows
representative networks for the
embryonic, larval, pupal, and adult stage of the development
respectively.}\label{fig:drosophila_nerv}
\end{figure}

We further studied the relations between 130 transcriptional
factors (TF). The network contains several clusters of transcriptional
cascades, and we will present the detail of the largest
transcriptional factor cascade involving 36 transcriptional
factors~(Figure~\ref{fig:drosophila_nerv}).  This cascade of TFs is
functionally very coherent, and many TFs in this network play
important roles in the nervous system and eye development.
For example,
Zn finger homeodomain 1 (zhf1),
brinker (brk),
charlatan (chn),
decapentaplegic (dpp),
invected (inv),
forkhead box, subgroup 0 (foxo),
Optix,
eagle (eg),
prospero (pros),
pointed (pnt),
thickveins (tkv),
extra macrochaetae (emc),
lilliputian (lilli), and
doublesex (dsx)
are all involved in nervous and eye development.  Besides functional
coherence, the network also reveals the dynamic nature of gene
regulation: some relations are persistent across the full
developmental cycle, while many others are transient and specific to
certain stages of development.  For instance, five transcriptional
factors, brk-pnt-zfh1-pros-dpp, form a long cascade of regulatory
relations which are active across the full developmental cycle.
Another example is gene Optix which is active across the full
developmental cycle and serves as
a hub for many other regulatory relations.  As for transience of the
regulatory relations, TFs to the right of the Optix hub reduced in their
activity as development proceeds to a later stage. Furthermore, Optix
connects two disjoint cascades of gene regulations to its left and
right side after embryonic stage.

\begin{figure}

\includegraphics{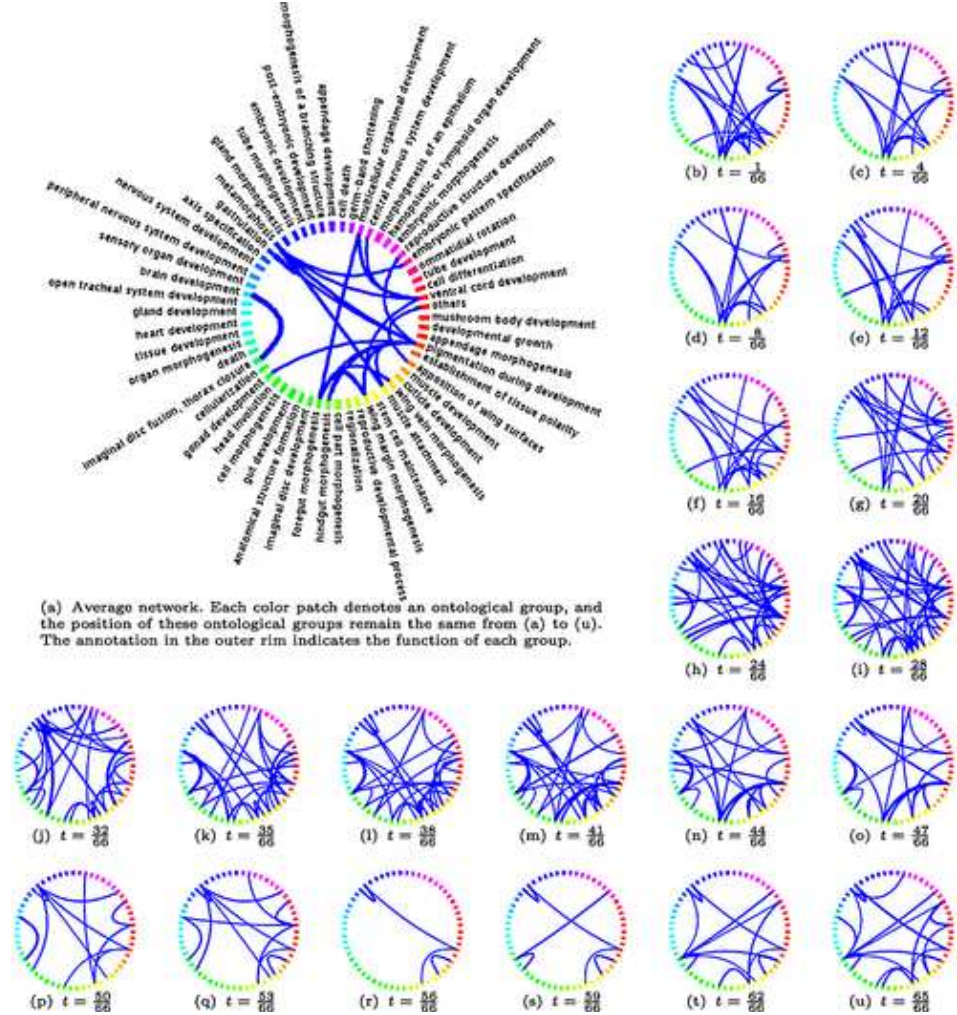}

\caption{Interactions between gene ontological groups related
to the developmental process undergo dynamic rewiring. The
weight of an edge between two ontological groups is the
total number of connections between genes in the two
groups. In the visualization, the width of an edge is
proportional to its edge weight. We thresholded the edge
weight at 30 in \textup{(b)--(u)} so that only those interactions
exceeding this number are displayed. The average network in
\textup{(a)} is produced by averaging the networks underlying
\textup{(b)--(u)}. In this case, the threshold is set to 20 instead.}
\label{fig:ont}
\end{figure}

The dynamic networks also provide an overview of the interactions
between genes from different functional groups.  In
Figure~\ref{fig:ont} we grouped genes according to 58 ontologies and
visualized the connectivity between groups. We can see that large
topological changes and network rewiring occur between functional
groups. Besides expected interactions, the figure also reveals many
seemingly unexpected interactions. For instance, during the transition
from pupa stage to adult stage, \textit{Drosophila} is undergoing a huge
metamorphosis. One major feature of this metamorphosis is the
development of the wing. As can be seen from
Figure~\ref{fig:ont}(r) and~(s),
genes related to metamorphosis, wing margin
morphogenesis, wing vein morphogenesis, and apposition of wing surfaces
are among the most active group of genes, and they carry their
activity into adult stage. Actually, many of these genes are also very
active during early embryonic stage [for example,
Figure~\ref{fig:ont}(b) and~(c)]; though the difference is they interact with different groups of genes. On one hand, the
abundance of the transcripts from these genes at embryonic stage is
likely due to maternal deposit~\citep{Arbeitmanetal2002}; on the other
hand, this can also be due to the diverse functionalities of these
genes. For instance, two genes related to wing development, held out
wings (how) and tolloid (td), also play roles in embryonic
development.

\section{Some properties of the algorithms} \label{sec:theory}

In this section we discuss some theoretical guarantees of the proposed
algorithms. The most challenging aspect in estimating time-varying
graphs is that the dimension of the data $p$ can be much larger than
the size of the sample $n$ ($p\gg n$), and there is usually only one
sample per time point. For example, in a genome-wide reverse
engineering task, the number of genes can be well over ten thousand
($p > 10{,}000$), while the total number of microarray measurements is
only in the hundreds ($n \sim 100$) and the measurements are collected at
different developmental stages. Then, the question is what are the
sufficient conditions under which our algorithms recover the sequence
of unknown graphs $\{ G^t \}_{t \in \mathcal{T}_n}$ correctly.

To provide asymptotic results, we will consider the model dimension
$p$ to be an increasing function of the sample size $n$, and
characterize the scaling of $p$ with respect to $n$ under which
structure recovery is possible. Furthermore, we will assume the following:
\begin{enumerate}
\item The graphs $\{ G^t \}_{t \in \mathcal{T}_n}$ are sparse, that is, the
maximum node degree $s$ of a graph is upper bounded and much smaller
than the sample size $n$ ($s \ll n$). The intuition here is that the
sparsity of a graph is positively correlated with the complexity of
a model; a sparse graph effectively limits the degree of freedom of
the model, which makes structure recovery possible given a small
sample size.
\item When regressing $X_u$ on $\Xv_{\setminus u}$, the relevant
covariates $S^\tau(u)$ should not be overly dependent on each
other. We need this assumption for the model to be
identifiable. Intuitively, if two covariates are very strongly
correlated with each other, we would not be able to distinguish one
from another.
\item The dependencies between irrelevant covariates $S^{\tau,c}(u)$
and those relevant ones $S^{\tau}(u)$ are not too strong. Similar to
the assumption 2, we need this assumption for the model to be
identifiable. Intuitively, if an irrelevant covariate looks very
similar to a relevant covariate, it will be hard for an algorithm to
tell which one is the true covariate. Assumptions 2 and 3 are common
in other work on sparse estimation, for example,
\citet{wainwright06sharp}; \citet{meinshausen06high}; \citet{ravikumar09high}.
\item The minimum parameter value $\theta_{\min}:= \min_{t \in \mathcal{
T}_n} \min_{v \in S^t(u)} |\theta_{uv}^t|$ is bounded away from
zero. This assumption is required in order to separate nonzero
parameters from zero parameters. If a covariate has very small
effect on the output, then it will be hard for the algorithm to
distinguish it from noise.
\item The parameter vector $\thetav^t$ is a smooth function of time.
This guarantees that the graphical models at adjacent time points
are similar enough such that we can borrow information across time
point by reweighting the observations. Under this assumption, the
method \verb|smooth| is able to achieve sufficiently fast
convergence rates for each neighborhood estimate; then the
consistency of the overall graph estimation can be achieved by an
application of the union bound over all nodes $u \in V$.
\item The kernel function $K(\cdot)$ is a symmetric nonnegative
function with a bounded support on $[-1,1]$. This assumption is
needed for technical reasons, and it gives some regularity
conditions on the kernel used to define the weights.
\end{enumerate}
With these assumptions, we can state the following theorem for
algorithm \verb|smooth| [a complete statement of assumptions 1--6 and
the proof of the theorem are given in~\citet{kolar09sparsistent}]:
\begin{theorem}[\citep{kolar09sparsistent}] \label{thm:main}
Assume that assumptions 1 to 6 given above hold. Let the regularization parameter
satisfy
\[
\lambda_1 \geq C  \frac{ \sqrt{\log p} }{n^{1/3}}
\]
for a constant $C > 0$ independent of $(n, p, s)$.  Furthermore,
assume that the following conditions hold:
\begin{enumerate}
\item $h = \mathcal{O}(n^{-1/3})$,
\item $s = o(n^{1/3})$, $\frac{s^3\log p}{n^{2/3}} = o(1)$,
\item $\theta_{\min} = \Omega(\frac{\sqrt{s\log p}}{n^{1/3}}).$
\end{enumerate}
Then for any $\tau \in \mathcal{T}_n$, the method \verb|smooth| estimates
a graph $\hat G^\tau$ that satisfies
\begin{equation}
\mathbb{P} [ \hat G^\tau \neq G^{\tau}  ] =
\mathcal{O}  \biggl( \exp  \biggl( -C\frac{n^{2/3}}{s^3} + C'\log p  \biggr)
\biggr) \rightarrow 0,
\end{equation}
for some constants $C', C''$ independent of $(n, p, s)$.
\end{theorem}

The theorem means that the procedure can recover the graph
asymptotically by using appropriate regularization parameter
$\lambda_1$, as long as both the model dimension $p$ and the maximum
node degree $s$ are not too large, and the minimum parameter value
$\theta_{\min}$ does not tend to zero too fast. In particular, the
model dimension is allowed to grow as $p = \mathcal{O}(\exp(n^\xi))$ for
some $\xi < 2/3$, when $s = \mathcal{O}(1)$ as is commonly assumed. The
consistency of the structure recovery is a somewhat surprising result
since at any time point there is at most one available sample
corresponding to each graph.

Currently we do not have a consistency result for the estimator
produced by the method \verb|TV|, however, we have obtained some
insight on how to solve this problem and plan to pursue it in our
future research. The main difficulty seems to be the presence of both
the $\ell_1$ and $\TV(\cdot)$ regularization terms in
equation~\eqref{eq:varying_coeff_tv}, which complicates the analysis.
However, if we relate the method \verb|TV| to the problem of multiple
change point detection, we can observe the following: the $\TV(\cdot)$
penalty biases the estimate $\{\hat{\thetav}{}^t\}_{t \in \mathcal{T}_n}$
toward a piecewise constant solution, and this effectively partitions
the time interval $[0,1]$ into segments within which the parameter is
constant. If we can estimate the partition $\mathcal{B}_u$ correctly,
then the graph structure can also be estimated successfully if there
are enough samples on each segment of the partition. In fact,
\citet{Rinaldo08properties} observed that it is useful to consider a
two-stage procedure in which the first stage uses the total variation
penalty to estimate the partition, and the second stage then uses the
$\ell_1$ penalty to determine nonzero parameters within each segment.
Although his analysis is restricted to the fused lasso
\citep{tibshirani05sparsity}, we believe that his techniques can be
extended for analyzing our method \verb|TV|. Besides assumptions 1 to
4 which appeared in method \verb|smooth|, additional assumptions may
be needed to assure the consistent estimation of the partition $\mathcal{
B}_u$.

\section{Discussion} \label{sec:discussion}

We have presented two algorithms for an important problem of structure
estimation of time-varying networks.  While the structure estimation
of the static networks is an important problem in itself, in certain
cases static structures are of limited use.  More specifically, a
static structure only shows connections and interactions that are
persistent throughout the whole time period and, therefore, time-varying structures are needed to describe dynamic interactions that
are transient in time. Although the algorithms presented in this paper
for learning time-varying networks are simple, they can already be
used to discover some patterns that would not be discovered using a
method that estimates static networks. However, the ability to
learn time-varying networks comes at a price of extra tuning
parameters: the bandwidth parameter $h$ or the penalty parameter
$\lambda_{\TV}$.

Throughout the paper, we assume that the observations at different
points in time are independent. An important future direction is the
analysis of the graph structure estimation from a general time series,
with dependent observations. In our opinion, this extension will be
straightforward but with great practical importance. Furthermore, we
have worked with the assumption that the data are binary, however,
extending the procedure to work with multi-category data is also
straightforward. One possible approach is explained in
\citet{ravikumar09high} and can be directly used here.

There are still ways to improve the methods presented here. For
instance, more principled ways of selecting tuning parameters are
definitely needed. Selecting the tuning parameters in the neighborhood
selection procedure for static graphs is not an easy problem, and
estimating time-varying graphs makes the problem more
challenging. Furthermore, methods presented here do not allow for the
incorporation of existing knowledge on the network topology into the
algorithm. In some cases, the data are very scarce and we would like
to incorporate as much prior knowledge as possible, so developing
Bayesian methods seems very important.

The method \verb|smooth| and the method \verb|TV| represent two
different ends of the spectrum: one algorithm is able to estimate
smoothly changing networks, while the other one is tailored toward
estimation of structural changes in the model. It is important to
bring the two methods together in the future work. There is a great
amount of work on nonparametric estimation of change points and it
would be interesting to incorporate those methods for estimating time-varying networks.

\section*{Acknowledgments}

We are grateful to two anonymous reviewers and the editor for their
valuable comments that have greatly helped improve the
manuscript.

\printaddresses

\end{document}